\documentclass[10pt,twocolumn,letterpaper]{article}
\usepackage[most]{tcolorbox}
\usepackage{multirow}
\usepackage{bbold}
\usepackage{tabularx}
\usepackage[table]{xcolor}

\usepackage[pagenumbers]{cvpr} 

\usepackage{ulem}

\definecolor{cvprblue}{rgb}{0.21,0.49,0.74}
\usepackage[pagebackref,breaklinks,colorlinks,allcolors=cvprblue]{hyperref}

\def\paperID{15441} 
\def\confName{CVPR}
\def\confYear{2026}

\title{Physion-Eval: Evaluating Physical Realism in Generated Video via Human Reasoning}

\author{
Qin Zhang$^{1}$\thanks{Corresponding author: qin@physionlabs.ai} \quad
Peiyu Jing$^{1}$ \quad
Hong-Xing Yu$^{2}$ \quad
Fangqiang Ding$^{3}$ \quad 
Fan Nie$^{2}$ \quad Weimin Wang$^{5}$\thanks{Work done outside of Character AI.} \\
Yilun Du$^{4}$ \quad
James Zou$^{2}$ \quad
Jiajun Wu$^{2}$ \quad
Bing Shuai$^{1}$ \\
\vspace{3pt}
$^{1}$Physion Labs \quad 
$^{2}$Stanford University \quad
$^{3}$MIT \quad $^{4}$Harvard University \quad $^{5}$Character AI \\
\vspace{4pt}
{\tt\small
\begin{tabular}{c}
\{qin,pyj,bing\}@physionlabs.ai \
\{koven,niefan,jamesz,jiajunwu\}@cs.stanford.edu  \\
wangweimin777@gmail.com \ 
\{fding\}@mit.edu \ ydu@seas.harvard.edu  
\end{tabular}
}
}

\begin{document}
\maketitle
\begin{abstract}
Video generation models are increasingly used as {world simulators} for storytelling, simulation, and embodied AI. As these models advance, a key question arises: {do generated videos obey the physical laws of the real world?} Existing evaluations largely rely on automated metrics or coarse human judgments such as preferences or rubric-based checks. While useful for assessing perceptual quality, these methods provide limited insight into when and why generated dynamics violate real-world physical constraints. We introduce {Physion-Eval}, a large-scale benchmark of expert human reasoning for diagnosing physical realism failures in videos generated by five state-of-the-art models across egocentric and exocentric views, containing {10,990} expert reasoning traces spanning {22} fine-grained physical categories. Each generated video is derived from a corresponding real-world reference video depicting a clear physical process, and annotated with temporally localized glitches, structured failure categories, and natural-language explanations of the violated physical behavior. Using this dataset, we reveal a striking limitation of current video generation models: in physics-critical scenarios, {83.3\%} of exocentric and {93.5\%} of egocentric generated videos exhibit at least one human-identifiable physical glitch. We hope Physion-Eval will set a new standard for physical realism evaluation and guide the development of physics-grounded video generation. The benchmark is publicly available at \href{https://huggingface.co/datasets/PhysionLabs/Physion-Eval}{huggingface.co/datasets/PhysionLabs/Physion-Eval}. 
\end{abstract}    
\begin{figure}[t!]
    \centering
    \includegraphics[width=0.96\linewidth]{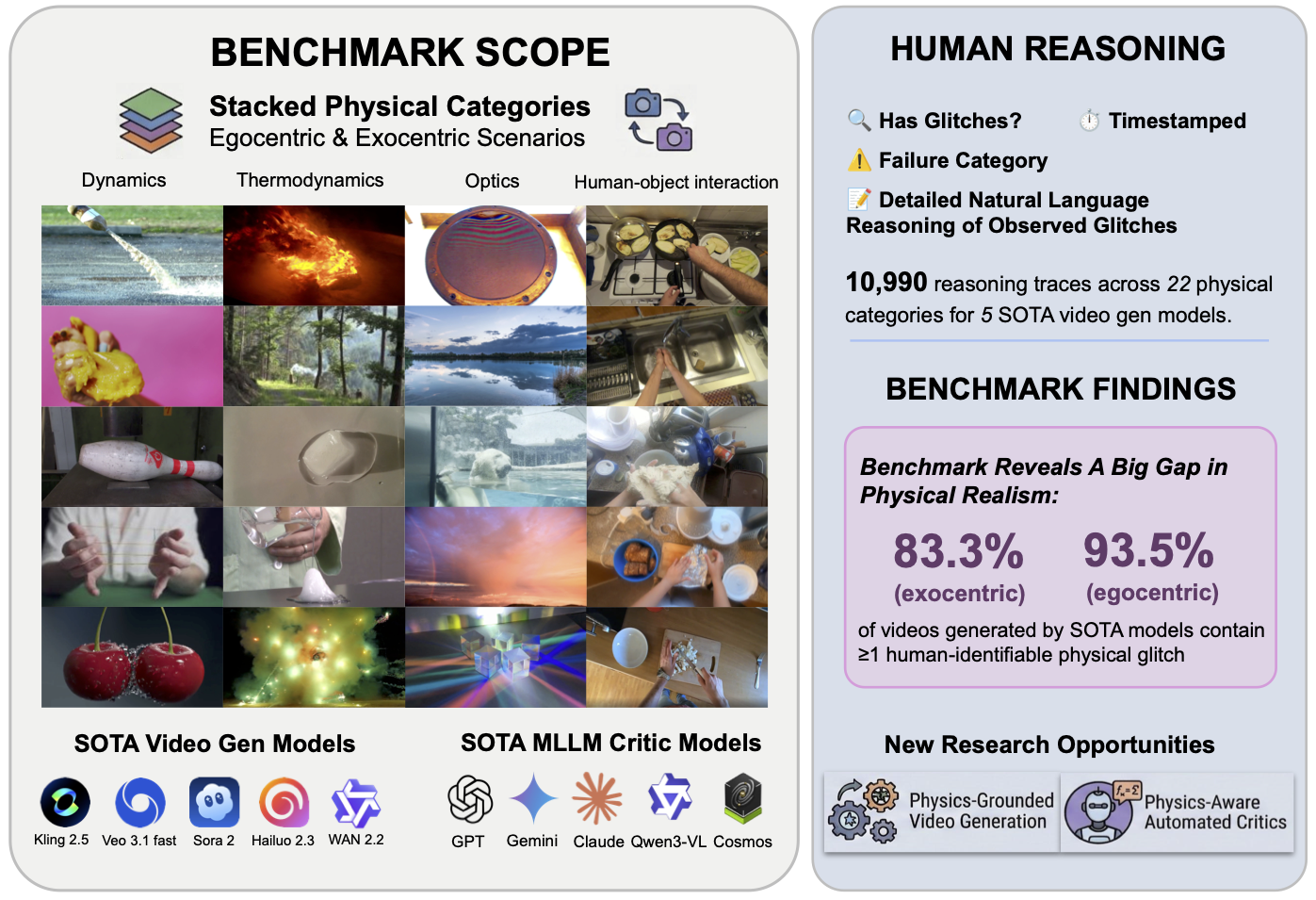}
    \caption{\textbf{Physion-Eval Benchmark.} \textit{(Left)} The benchmark spans diverse physical phenomena across egocentric and exocentric views, evaluating videos generated by five state-of-the-art generation models. \textit{(Right, top)} {Physion-Eval} provides {10,990} expert-annotated reasoning traces with timestamped glitch localization, structured failure categories, and natural-language explanations. \textit{(Right, bottom)} Results reveal a large physical realism gap: {83.3\%} of exocentric and {93.5\%} of egocentric generated videos contain at least one human-identifiable physical glitch, motivating physics-grounded video generation and automated critics.
    }
    \label{fig:first}
\end{figure}

\section{Introduction}
\label{sec:intro}
Video generation models are rapidly evolving from tools for visual synthesis into systems capable of simulating dynamic physical worlds. Recent models such as Veo 3.1~\cite{google2025veo31} and Sora 2~\cite{openai2025sora2} can generate scenes with increasingly coherent lighting, materials, motion, and articulated behavior~\cite{brooks2024video,bansal2024videophy,agarwal2025cosmos,jang2025dreamgen,ParkerHolder2025Genie3,zhou2025hermes}. As such capabilities advance, video generation is emerging as a new computational medium for modeling and interacting with reality, with applications spanning filmmaking, advertising, interactive simulation, and embodied AI. In this setting, visual plausibility alone is insufficient: generated videos must also respect the physical principles governing the real worlds they depict, where objects persist over time, forces produce plausible outcomes, and events unfold with coherent causal structure.

\begin{figure*}[t!]
    \centering\includegraphics[width=0.91\linewidth]{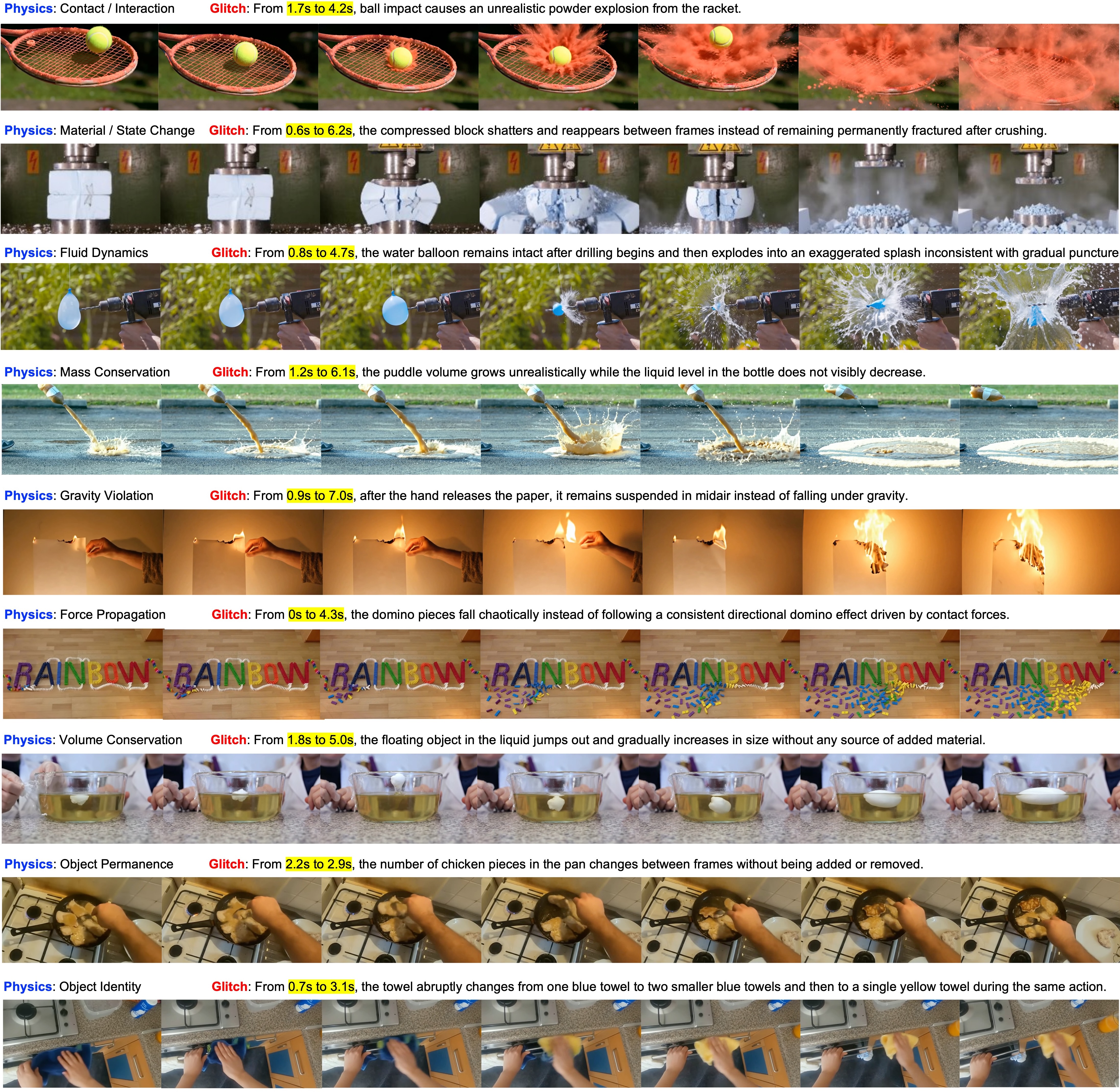}
    \caption{\textbf{Examples of physical glitches in AI-generated videos from Physion-Eval.} Each row shows a representative failure mode where generated dynamics violate basic physical principles. Frame sequences illustrate how these glitches emerge over time.}
    \label{gallery}
\end{figure*}

Despite impressive progress in generation quality, evaluating physical realism in video generation remains an open challenge. Prior benchmarks such as VideoPhy~\cite{bansal2024videophy,bansal2025videophy}, PhyGenBench~\cite{meng2024towards}, Cosmos-Eval~\cite{anonymous2025cosmoseval} and PhyWorldBench~\cite{gu2025phyworldbench} have begun exploring this problem. However, most prior work relies on automated metrics or model-based critics to assess compliance with a limited set of physical scenarios~\cite{bansal2024videophy,meng2024towards}. While useful, these signals often correlate weakly with human judgments~\cite{anonymous2025cosmoseval} and struggle to detect subtle violations of dynamics, contact, and causality that require precise spatial and temporal reasoning. Additionally, existing evaluations primarily focus on exocentric viewpoints~\cite{gibson2014ecological,luo2024put,he2025bridging}, leaving egocentric scenarios largely unexplored, despite their importance for immersive media and embodied AI systems where physical consistency is critical.

In this work, we introduce \textit{\textbf{Physion-Eval}}, a new benchmark for evaluating perceptual physical realism in video generation, grounded in large-scale expert human reasoning. Physion-Eval is built from WISA-80K~\cite{wang2025wisa} and EPIC-KITCHENS~\cite{damen2020epic}, and covers 22 fundamental physical phenomena\footnote{The 22 categories include 17 exocentric phenomena from WISA~\cite{wang2025wisa} (6 dynamics, 6 thermodynamics, and 5 optics) and 5 egocentric physical interaction categories defined in~\cref{tab:epic_physics_verbs}.} across both exocentric and egocentric viewpoints. The benchmark is curated to include videos with clear, visually observable physical interactions, providing strong signals for evaluating physical realism in AI-generated videos. To support fine-grained analysis, Physion-Eval includes 10,990 expert-annotated reasoning traces collected from ninety expert annotators, each with precise temporal localization of physical glitches and natural-language explanations of the underlying violated principles, spanning outputs from five state-of-the-art video generation models. Using this benchmark, we uncover a striking physical realism gap: {{83.3\% of exocentric and 93.5\% of egocentric videos generated by leading video generation models contain at least one human-identifiable violation}} of contact, force, timing, or causality. A gallery of example physical realism failures is shown in~\cref{gallery}. These results suggest that modeling real-world physical dynamics remains a significant challenge for current video generation systems. 

We further examine whether state-of-the-art MLLM critics can detect and reason about those physical realism failures identified by humans. To this end, we evaluate 10 proprietary and open-source MLLMs, including models from the Gemini family~\cite{google2025gemini25,google2025gemini3provision}, QWEN3-VL~\cite{bai2025qwen3vl}, and Cosmos-Reason~\cite{agarwal2025cosmos,nvidia2026cosmosreason2docs}, among others. Across both egocentric and exocentric settings, we observe a consistent and substantial performance gap between MLLM critics and human observers. As shown in~\cref{fig:teaser}, Gemini 3.0 Pro~\cite{google2025gemini3provision} fails to identify over 74.4\% of exocentric and 90.1\% of egocentric videos that contain clearly visible glitches that untrained viewers can readily detect. Qualitative analysis in \cref{fig:reasoning_trace} and Appendix \cref{fig:MLLM_reasoning_vs_human_reasoning} reveals systematic discrepancies between MLLM and human reasoning. MLLMs frequently produce hallucinated explanations and incorrect temporal localization, and often fail to accurately determine when and why physical inconsistencies occur, particularly for violations that unfold over time. These observations suggest that current MLLM critics struggle with temporally grounded reasoning and reliable causal attribution in physical processes.

This work makes three key contributions. First, we curate a large-scale, physics-rich video dataset spanning diverse physics scenarios and conduct a human study with ordinary viewers, who represent the typical audience of generated media. We show that state-of-the-art video generation models frequently produce physical glitches that are readily detectable by such viewers in both exocentric and egocentric settings. Second, we benchmark leading MLLM critics and find that they largely fail to detect these glitches, revealing a large gap between human perception and automated evaluation. Last, we introduce Physion-Eval, an expert-annotated benchmark for perceptual physical realism containing 10,990 reasoning traces, with timestamped glitch localization, structured failure categories, and natural-language explanations. Expert annotation enables consistent taxonomy-level labeling and temporally grounded diagnostic reasoning beyond simple preferences. To our knowledge, Physion-Eval is the first and largest dataset of temporally grounded human reasoning annotations for diagnosing physical realism failures in generated videos. We hope it will facilitate the development of more reliable, physically grounded video generation.
\section{Related Work}
\label{sec:related_works}

\noindent\textbf{Video Generation Model.} \ Video generation models learn a conditional distribution over videos, $p(x_{1:T}|c)$, where $x_{1:T}$ are frames within a $T$-duration video, and the conditioning $c$ can be text, images or videos. Modern models typically use transformer backbones\cite{peebles2022dit,brooks2024video,teamwan2025wan,yang2024cogvideox,ma2024latte,lin2024opensoraplan,opensora2025opensora20,kong2024hunyuanvideo,wu2025hunyuanvideo15,agarwal2025cosmos,nvidia2025cosmospredict25,alhaija2025cosmostransfer1}, and the distribution is modeled in a compressed spatial-temporal latent space produced by a video VAE (or a learned tokenizer), rather than directly in the pixel space
\cite{blattmann2023svd,chen2024videocrafter2,zheng2024opensora,yang2024cogvideox,kong2024hunyuanvideo,wu2025hunyuanvideo15,teamwan2025wan,agarwal2025cosmos,nvidia2025cosmospredict25,zhao2024cvvae,wu2024ivvae,yin2025decovae}.
The model is often trained with a diffusion denoising objective \cite{peebles2022dit,song2021sde,lipman2022flowmatching,liu2022rectifiedflow} from large-scale real-world videos.
This learning paradigm underpins state-of-the-art work \cite{brooks2024video,teamwan2025wan,opensora2025opensora20,nvidia2025cosmospredict25} and commercial video generation models. Recent works \cite{brooks2024video,teamwan2025wan,opensora2025opensora20,kong2024hunyuanvideo,wu2025hunyuanvideo15,nvidia2025cosmospredict25,agarwal2025cosmos} show phenomenal progress in generating videos with improved motion continuity, camera dynamics, and prompt adherence.  However, as we show, these advances do not reliably translate into improved {perceptual physical realism}. Generated videos still exhibit implausible contact, wrong force responses, and violations of basic physical constraints. Several aspects of the prevailing training paradigm likely underlie this problem: the denoising objectives \cite{peebles2022dit,song2021sde,lipman2022flowmatching,liu2022rectifiedflow} reward appearance-consistent reconstruction in latent space rather than enforcing physical constraints, and Internet-scale video corpora overrepresent common motions and cinematic edits while underrepresenting clean, constraint-revealing physical interactions, which biases models toward visual aesthetics over physical correctness \cite{blattmann2023svd,chen2024videocrafter2}. Closing this gap is important for deploying video generation models in real-world physical AI applications.

\begin{table*}[t!]
\caption{Mapping from EPIC-KITCHENS~\cite{damen2018scaling} verb labels to visually observable physical categories used in the egocentric setting. }
\vspace{-3pt}
\footnotesize
\centering
\setlength{\tabcolsep}{4pt}
\renewcommand{\arraystretch}{1.2}
\begin{tabular}{p{0.2\textwidth} p{0.25\textwidth} p{0.49\textwidth}}
\toprule
\textbf{Physical Category} & \textbf{Visually Observable Physics} & \textbf{Mapped EPIC-KITCHENS Action Verbs} \\
\toprule

\textbf{Rigid-Body Interaction} 
& \scriptsize{Rigid motion, contact, articulation, gravity, momentum}
& \scriptsize{take, put, hold, carry, move, lift, remove, drop, let-go, set, open, close, turn, turn-on, turn-off, unlock, lock, press, push, pull, switch, adjust, use, attach, detach, throw} \\
\hline

\textbf{Deformation \& Fracture} 
& \scriptsize{Elastic deformation, cutting, separation}
& \scriptsize{bend, flatten, wrap, unwrap, unroll, cut, break, crush, stab, divide, mark, score, sharpen} \\
\hline

\textbf{Soft Materials \& Mixing} 
& \scriptsize{Cloth dynamics, viscoplastic flow}
& \scriptsize{mix, stir, shake, knead, wear, dry, fold} \\
\hline

\textbf{Fluid \& Granular Flow} 
& \scriptsize{Liquid flow, particle scattering and piling}
& \scriptsize{pour, fill, empty, wash, water, soak, spray, sprinkle, season, grate} \\
\hline

\textbf{Thermal \& Frictional Effects} 
& \scriptsize{Heating/cooling cues, friction, surface wear}
& \scriptsize{cook, bake, unfreeze, scrub, brush, rub} \\
\bottomrule
\end{tabular}
\label{tab:epic_physics_verbs}
\end{table*}

\vspace{3pt}
\noindent\textbf{Evaluation of Physics in Vide Generation.} \ Video generation quality is usually evaluated via distributional realism using FVD~\cite{unterthiner2018fvd}
, motion and temporal coherence using FVMD~\cite{liu2024fvmd}, prompt alignment using CLIP-based similarity~\cite{radford2021clip,hessel2021clipscore}, and reference-based fidelity when ground truth exists using LPIPS~\cite{zhang2018lpips} and SSIM~\cite{wang2004ssim}, with multi-attribute benchmark suites such as VBench~\cite{huang2024vbench} and VBench++~\cite{huang2024vbenchpp}. In recognizing the lack of reliable measures for physical realism in generated videos, an emerging body of literature \cite{motamed2025physicsiq,guo2025t2vphysbench,zhang2025morpheus,bansal2025videophy,meng2024towards,gu2025phyworldbench,duan2025worldscore}  propose various metrics and benchmark to fill this gap. Physics-IQ \cite{motamed2025physicsiq} and Morpheus \cite{zhang2025morpheus} quantify physical plausibility by extracting statistics from salient object trajectories and interactions, but such object-centric formulations do not extend to non-object-dominated phenomena such as fluid dynamics or combustion. WorldScore \cite{duan2025worldscore} measures 3D consistency and photometric consistency, yet it is limited to static scenes. In parallel, PhyGenEval \cite{meng2024towards}, PhyWorldBench \cite{gu2025phyworldbench}, PhysBench \cite{chow2025physbench}, and VideoPhy-2 \cite{motamed2025physicsiq} adopt zero-shot MLLM judges for physical realism evaluation, but we show in~\cref{untrained_results} that this approach can be unreliable. Finally, existing evaluations are limited in scale and coverage of their prompts and datasets, whereas we present a large-scale human evaluation spanning state-of-the-art closed- and open-source models.

\section{Video Source Curation}\label{sec:video_source}
Existing video generation benchmarks lack diversity in physical scenes and viewpoints. To address this, we curate a dataset of generated videos that span diverse physical processes, grounded in real-world videos across both exocentric and egocentric views. Each generated video is conditioned on a real video’s first frame and video caption, and is designed to focus on a single observable physical interaction while minimizing confounding factors.

\vspace{3pt}
\noindent \textbf{Exocentric Videos.} \ We source exocentric videos from WISA-80K~\cite{wang2025wisa}, which is a large-scale natural video dataset covering 17 fundamental physical phenomena across dynamics, thermodynamics, and optics. Each video is paired with a caption (provided by WISA) and a human-assigned physics category label. We perform manual filtering to remove low-quality samples, including multi-shot clips, temporally reversed videos, near-duplicates, videos with overlays or synthetic content, and samples with incorrect physics labels. We further balance the dataset to obtain near-uniform coverage across all physical categories, resulting in 1,734 curated videos. Detailed filtering and cleaning procedures are provided in Appendix~\cref{sec:wisa_cleaning}.

\vspace{3pt}
\noindent\textbf{Egocentric Videos.} \ EPIC-KITCHENS~\cite{damen2018scaling} is a large-scale egocentric video dataset with fine-grained action annotations and precise temporal boundaries. We construct an egocentric short-clip dataset of 752 videos by extracting 4-second to 9-second action segments using its provided verb-labeled action timestamps. Each verb is mapped to a physical category aligned with the WISA taxonomy (Table~\ref{tab:epic_physics_verbs})\footnote{While EPIC-KITCHENS contains a broader set of verbs, we exclude meta-level verbs (\textit{transition}, \textit{prepare}, \textit{finish}) that denote process boundaries, as well as sensory actions (\textit{look}, \textit{feel}, \textit{smell}, \textit{wait}), which do not involve meaningful physical interactions.}. These clips capture concrete physical interactions (e.g., cutting, pouring, grasping) from an egocentric viewpoint, closely mirroring embodied agent interactions in physical AI systems~\cite{runway_gwm1_2025,1x_world_model,ali2025humanoid}. For each extracted video, we generate a caption using Gemini 2.5 Pro~\cite{google2025gemini25}, conditioned on the action verb (see Appendix~\cref{prompt} for the prompt), and have humans manually review for accuracy.

\vspace{3pt}
\noindent \textbf{Construction of Generated Videos.} \ 
For each real-world video, we use its video caption and the first visually non-black video frame\footnote{We define the first visually non-black frame as the earliest frame where, after denoising and masking constant mattes, either $\geq4\%$ of pixels have HSV$\geq28$ or $\geq1\%$ have saturation $\geq35$ and value $\geq23$, excluding fade-to-black slates and sensor noise while capturing real visual content.} as conditioning inputs, and prompt latest text-and-image-to-video (TI2V) models, including Sora 2~\cite{openai2025sora2}, Veo 3.1 fast~\cite{google2025veo31}, Kling 2.5~\cite{klingai2025kling25}, Hailuo 2.3~\cite{minimax2025hailuo23} and Wan 2.2~\cite{teamwan2025wan}, to synthesize video twins. Because TI2V models produce outputs with varying resolutions and durations, we standardize all videos by center-cropping to a 16:9 aspect ratio and resizing to 720×1280. Moreover, as our evaluation focuses on visually perceptible physical glitches, we remove the audio from all videos. In total, we generate 12,718 videos from 2,486 real-world source videos, all processed using the same standardization pipeline.

\begin{figure*}[t!]
    \centering
    \includegraphics[width=1\linewidth]{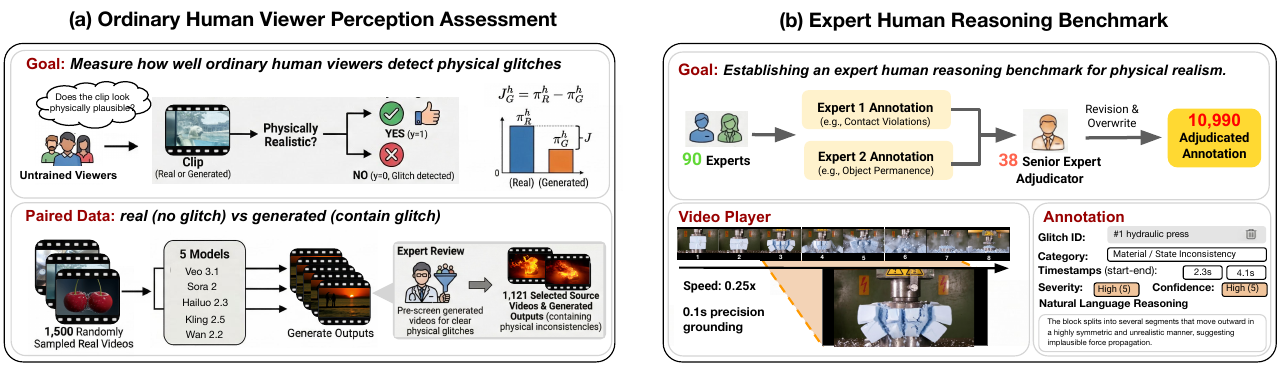}
    \caption{Two complementary human evaluation studies for assessing physical realism in generated videos. \textbf{(a) Perceptual detection by ordinary viewers.} Untrained viewers evaluate a blinded 1:1 mixture of real-world videos and outputs from five video generation models, judging whether each clip appears physically realistic. The evaluation metric measures how often generated videos are perceived as physically realistic relative to real videos.
\textbf{(b) Physion-Eval expert reasoning benchmark.} Expert annotators follow a three-expert workflow to annotate generated videos, producing temporally localized failures, category labels, severity scores, and natural-language explanations. The final dataset contains 10,990 adjudicated reasoning annotations for diagnosing failure modes in video generation models.}
    \label{fig:expert_annotation}
\end{figure*}

\section{Human Evaluation of Physical Realism}
To capture both perceptual judgments and diagnostic understanding of physical realism in generated videos, we conduct two complementary human studies. First, we measure perceptual detectability using untrained viewers, reflecting the typical audience of generated media, to assess whether physical implausibilities are noticeable by ordinary people. Second, we introduce an expert annotation benchmark that provides temporally grounded, taxonomy-based diagnoses with structured explanations of violated physical principles, enabling systematic diagnosis of video generation models.

\subsection{Perceptual Detection by Ordinary Viewers}\label{sec:human_eval}
\subsubsection{Experiment Setup}
\noindent\textbf{Study Design.} \ 
We recruit 16 untrained viewers with no affiliation with the authors. To construct the evaluation set, we randomly sample 1,500 source videos from~\cref{sec:video_source} and generate outputs using all five TI2V models, and identify 1,121 sources whose generated videos exhibit clear physical glitches. We further trim generated videos to 4–8 seconds and remove the first 20 frames from all clips to eliminate duration cues and initialization artifacts~\cite{ren2024consisti2v,wang2024generative}. For each model, viewers evaluate 100 exocentric and 100 egocentric videos from a randomly ordered 1:1 mix of real and generated clips. They judge whether each clip appears physically realistic based solely on visual evidence, labeling it realistic if no clear glitch is observed. To avoid bias, real and generated videos from the same source clip are never shown to the same viewer. This design prioritizes recall of physical realism by avoiding penalties for ambiguous but plausible dynamics, making the results conservative upper bounds on perceived physical realism. In total, we collect over 12,000 judgments from untrained viewers.

\begin{figure*}[t!]
    \centering
    \includegraphics[width=0.95\linewidth]{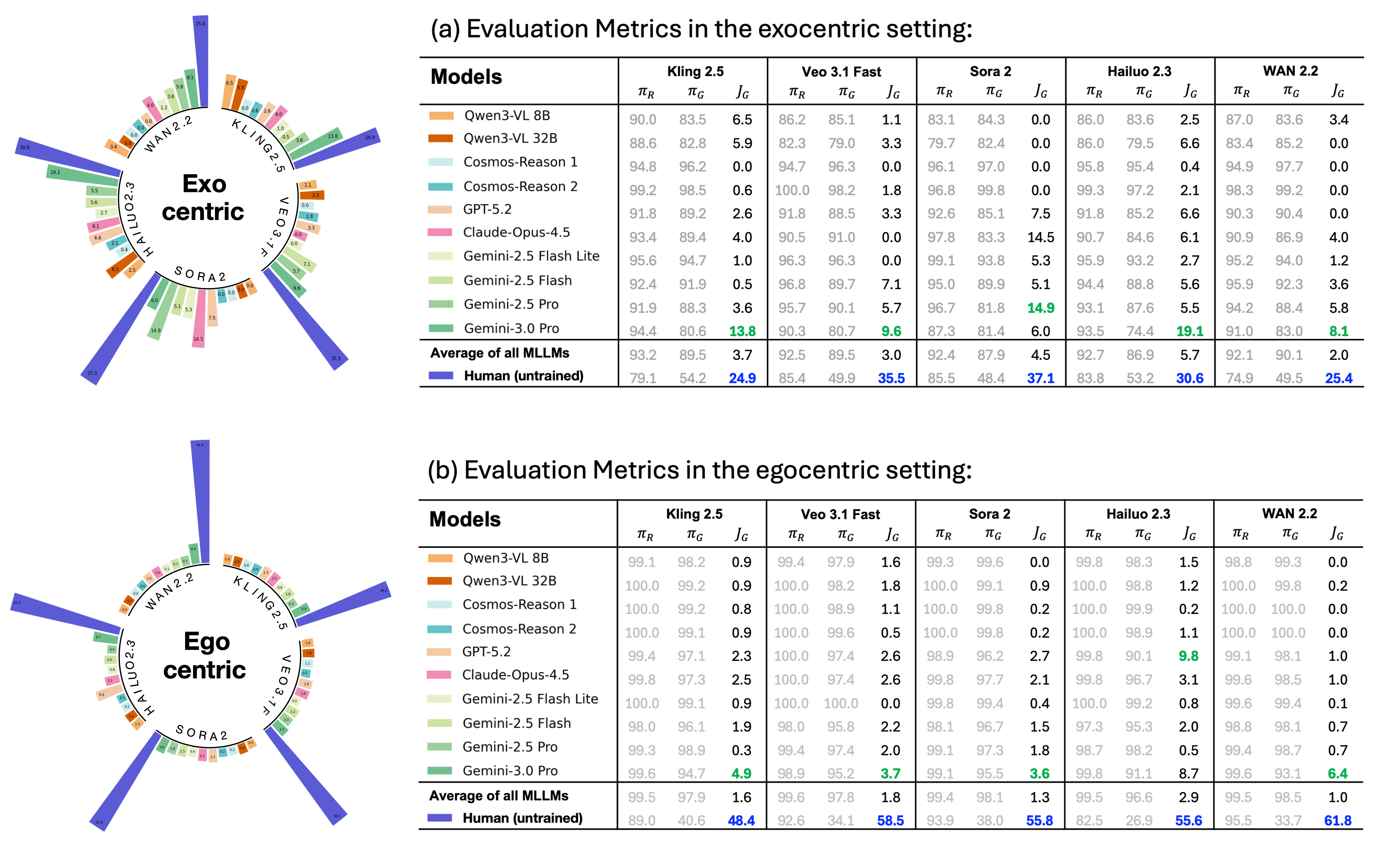}
    \caption{
        \textbf{Evaluation results of the untrained human study across video generation models under (a) exocentric and (b) egocentric settings.} The radial plots (left) visualize Youden's J statistic ($J_G$) for each evaluator, while the tables (right) report the corresponding metrics $\pi_R$, $\pi_G$ and $J_G$. Across models, untrained human viewers consistently achieve higher J scores than current MLLM critics, indicating a stronger sensitivity to physical glitches in genereated videos, especially in the egocentric setting.
    }
    \label{fig:teaser}
\end{figure*}

\vspace{3pt}
\noindent\textbf{Evaluation Metrics.} \ Because generated videos are pre-screened to contain clear physical glitches while real videos are glitch-free, we can compute how often each is judged physically realistic. To quantify the difference in how often real and generated videos are judged as physically realistic, we adopt Youden’s J statistic~\cite{youden1950index}. Let $P_R$ and $P_G$ denote the distributions of real and generated videos, respectively. Given binary judgments where $y_h(v)=1$ indicates evaluator $h$ judges video $v$ as physically realistic and $y_h(v)=0$ otherwise, we define the average perceived physical realism of real and generated videos as:
\begin{equation}
\pi_R^h = \underset{v \sim P_R}{\mathbb{E}}\big[\mathbb{E}_{h}[y_h(v)]\big], \ \ \ \pi_G^h = \underset{v \sim P_G}{\mathbb{E}}\big[\mathbb{E}_{h}[y_h(v)]\big]
\end{equation}
where higher values of $\pi$ indicate stronger physical realism. We then adopt the J statistic~\cite{youden1950index} under our definition as:
\begin{equation}
\label{j_def}
J_G^h = \pi_R^h - \pi_G^h
\end{equation}
The \(J\) statistic measures the drop in perceived physical realism from real to generated videos. Unlike deepfake detection~\cite{rana2022deepfake,shen2025authguard}, which focuses on real-synthetic classification, \(J\) evaluates semantic physical realism in an origin-agnostic setting where both real and generated videos may be physically plausible or implausible (see Appendix~\cref{defition} for detailed definition). A value of \(J=0\) indicates that generated and real videos are judged realistic at the same rate (negative values are clipped to zero). Larger \(J\) values indicate greater perceptual degradation due to physical glitches in generated videos.

\vspace{3pt}
\noindent\textbf{MLLM Critic Evaluation.} \ We also evaluate state-of-the-art MLLM critics to assess how well automated evaluators align with the judgments of ordinary human viewers about physical realism. We benchmark 10 models, including GPT-5.2~\cite{openai2025gpt52}, Gemini (3.0 Pro, 2.5 Pro, 2.5 Flash, 2.5 Flash Lite)~\cite{google2025gemini25,google2025gemini3provision}, Claude-4.5 Opus~\cite{anthropic2025claudeopus45}, Qwen-3-VL-8B / 32B~\cite{bai2025qwen3vl}, and Cosmos Reason 1 / 2~\cite{nvidia2025cosmosreason1physicalcommonsense,nvidia2026cosmosreason2docs}. All models use the same prompt (see Appendix~\cref{prompt}).

\subsubsection{Results}\label{untrained_results}
In \cref{fig:teaser}, we report \(J\) computed from 12,000 judgments by untrained viewers. As shown, untrained humans reliably detect physical inconsistencies, achieving \(J=24.9\%\text{--}37.1\%\) (exocentric) and \(48.4\%\text{--}61.8\%\) (egocentric). We hypothesize that untrained viewers detect more glitches in egocentric videos due to limited egocentric training data and stronger camera motion, which introduces more challenging viewpoint dynamics for current video generation models. In contrast, the best MLLM critics reach only \(J=19.1\%\) (exocentric) and \(9.8\%\) (egocentric), respectively. The gap is larger in egocentric videos, indicating that first-person views make physical violations more perceptually salient. Across the evaluated MLLM critics, we observe $\pi_G \to 1$, indicating that MLLMs frequently judge generated videos as physically realistic even when clear violations are present (e.g., objects passing through each other or motion reversing without cause). Due to space constraints, we defer ablation studies of critic performance, such as temporal sampling and extended reasoning (“thinking”), to Appendix~\cref{temporal_sampling,thinking}, where we find these techniques offer little improvement for physical realism detection. We also defer analysis of how physical intensity and dynamics affect human and MLLM perception of physical realism to Appendix~\cref{sec:physcial_dependency}. There, we show that human judgments are sensitive to these factors, while MLLMs fail to detect glitches regardless of the intensity or dynamics of the physical process.
\subsection{Physion-Eval: Physical Reasoning Benchmark}

\begin{figure*}[t!]
    \centering
    \includegraphics[width=1\linewidth]{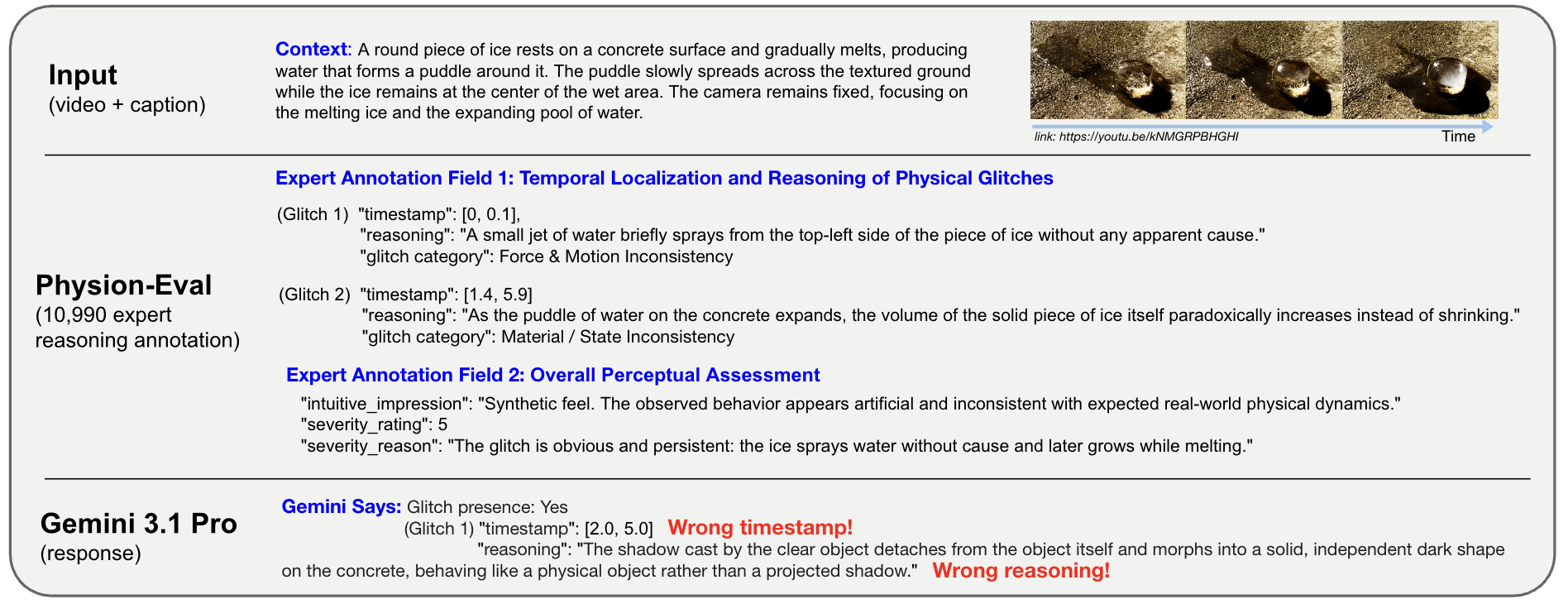}
    \caption{\textbf{Example from Physion-Eval comparing expert human annotations and MLLM reasoning.} In this example, human annotators correctly identify that the ice object sprays water without a visible cause and later increases in volume while melting, violating expected causal behavior and mass conservation. In contrast, Gemini 3.1 Pro hallucinates a non-existent shadow artifact, highlighting a substantial gap between human reasoning and current automated critics.}
    \label{fig:reasoning_trace}
\end{figure*} 

While \cref{sec:human_eval} shows that untrained viewers can readily detect physical glitches far better than current MLLMs, converting these failures into diagnostic signals requires domain expertise. To this end, we introduce \textbf{Physion-Eval}, a large-scale expert reasoning dataset for physical realism violations in generated videos. The dataset contains 12,718 generated videos and 10,990 expert reasoning traces with temporally grounded failure localization, structured glitch categories, and natural-language explanations.

\subsubsection{Annotation Protocol}
We recruit 90 expert annotators with bachelor’s degrees in STEM fields (e.g., physics or engineering) and formal training in undergraduate physics. All expert annotators complete six training sessions based on detailed guidelines (Appendix~\cref{guideline}), with quality monitored via cross-annotator agreement and  similarity to ground-truth annotations. The top performers are promoted to senior experts, resulting in 38 senior annotators. During the main annotation phase (\cref{fig:expert_annotation}(b)),  two experts independently annotate each video, and a third senior annotator reviews both annotations from the previous two annotators and adjudicates disagreements to produce the final annotation. Each final annotation contains four components:
\begin{enumerate}
\item \textit{Glitch Presence (True/False)}: Determination of whether the video contains any physical inconsistency.
\item \textit{Temporal Grounding}: Timestamp localization of each failure with 0.1-second precision.
\item \textit{Glitch Classification}: Assignment of a failure category from a predefined taxonomy, including categories such as 1) contact/interaction failures, 2) object permanence violations, 3) temporal coherence breakdowns, 4) causal sequence violations, 5) force and motion inconsistencies, 6) material/state inconsistencies, 7) geometric/collision violations, and others. Detailed definition for each category can be found in Appendix~\cref{guideline}.
\item \textit{Reasoning}: A natural-language explanation describing the violated physical behavior.
\end{enumerate}

To support this, we design a custom annotation workflow and user interface (\cref{fig:expert_annotation}(b)) that enforces a taxonomy-first protocol. Expert annotators first assign a failure category, then provide detailed, temporally grounded annotations for each identified glitch. Multiple anomalies within a video are recorded as separate instances, each with its own timestamp, category label, and supporting evidence. The interface supports slow-motion playback and fine-grained temporal selection for precise localization. A dedicated review mode enables senior experts to inspect, revise, and finalize annotations, producing consistent, taxonomy-aligned reasoning traces for systematic analysis.

\subsubsection{Comparison of Human and MLLM Reasoning}
\cref{fig:reasoning_trace} compares expert human and MLLM reasoning on physical realism. In this example, human annotators identify two failures with precise timestamps: (1) an uncaused water spray and (2) the ice increasing in volume while melting, both grounded in timestamps. In contrast, existing MLLM critics fail catastrophically, especially on reasoning when failures occur and on eliciting the correct reasons. For example, Gemini 3.1 Pro produces an incorrect timestamp and attributes the failure to a shadow artifact, which reflects a hallucination. This pattern is not limited to isolated cases but is consistent across examples (Appendix~\cref{comparison_mllm_vs_human}): MLLM critics often mislocalize failures in time and hallucinate causal explanations, while humans provide accurate, temporally grounded reasoning. This suggests that human judgment remains the gold standard for evaluating physical realism in generated videos.

\begin{table*}[t]
\centering
\small
\setlength{\tabcolsep}{3pt}
\begin{tabular}{l|cccc|cccc}
\toprule 
& \multicolumn{4}{c|}{\textbf{Exocentric}} & \multicolumn{4}{c}{\textbf{Egocentric}} \\
\cmidrule(lr){2-5} \cmidrule(lr){6-9}
\textbf{Model} & \textbf{\# Videos} & \textbf{Failure Rate ($\downarrow$)} & \textbf{Density ($\downarrow$)} & \textbf{Severity ($\downarrow$)} & \textbf{\# Videos} & \textbf{Failure Rate ($\downarrow$)} & \textbf{Density ($\downarrow$)} & \textbf{Severity ($\downarrow$)} \\

\midrule
Kling 2.5~\cite{klingai2025kling25}     
& 1,738 & \textcolor{ForestGreen}{\textbf{73.8\%}} & \textcolor{ForestGreen}{\textbf{1.15$\pm$1.08}} & \textcolor{ForestGreen}{\textbf{2.69$\pm$1.80}} & 416 & 96.4\% & 1.42$\pm$1.06
& 3.05$\pm$1.58 \\

Veo3.1 Fast~\cite{deepmind2025veo}   
& 1,696 & 79.4\%  & 1.32$\pm$1.11 
& 3.01$\pm$1.74 & 402 & 97.5\% & 1.69$\pm$1.12 & 3.37$\pm$1.56 \\

Sora 2~\cite{opensora2025opensora20}        
& 1,587 & 79.2\%  & 1.21$\pm$0.94 
& 2.88$\pm$1.67 & 763 & 96.6\% & \textcolor{ForestGreen}{\textbf{1.23$\pm$0.93}} & \textcolor{ForestGreen}{\textbf{2.81$\pm$1.69}} \\

Hailuo 2.3~\cite{minimax2025hailuo23}  & 1,719 & 93.1\%  
& {1.42$\pm$0.90} 
& {3.61$\pm$1.62} & 423 & 92.0\% & {1.92$\pm$1.36} & {3.86$\pm$0.88} \\

Wan 2.2~\cite{teamwan2025wan}  & 1,751 & 90.3\% 
& 1.32$\pm$0.88 
& 3.33$\pm$1.39 & 449 & \textcolor{ForestGreen}{\textbf{83.5\%}} & 1.56$\pm$1.36 & 3.49$\pm$0.75 \\

\midrule
\rowcolor{gray!20}
Average & 1,707.4 &	83.3\% &	1.28$\pm$0.98	& 3.10$\pm$1.64 & 490.6 & 	93.5\%	& 1.56$\pm$1.17	& 3.32$\pm$1.29\\
\bottomrule
\end{tabular}
\caption{\textbf{Physical glitch statistics across video generation models.} \textit{Failure rate} denotes the percentage of generated videos that contain at least one human-identified glitch. \textit{Glitch density} denotes the average number of glitches per video. \textit{Glitch severity} denotes the average glitch severity score reported by the expert annotators. We also report mean $\pm$ standard deviation for glitch density and severity.}
\label{tab:glitch_metrics}
\end{table*}

\begin{figure}[t!]
    \centering
    \includegraphics[width=1\linewidth]{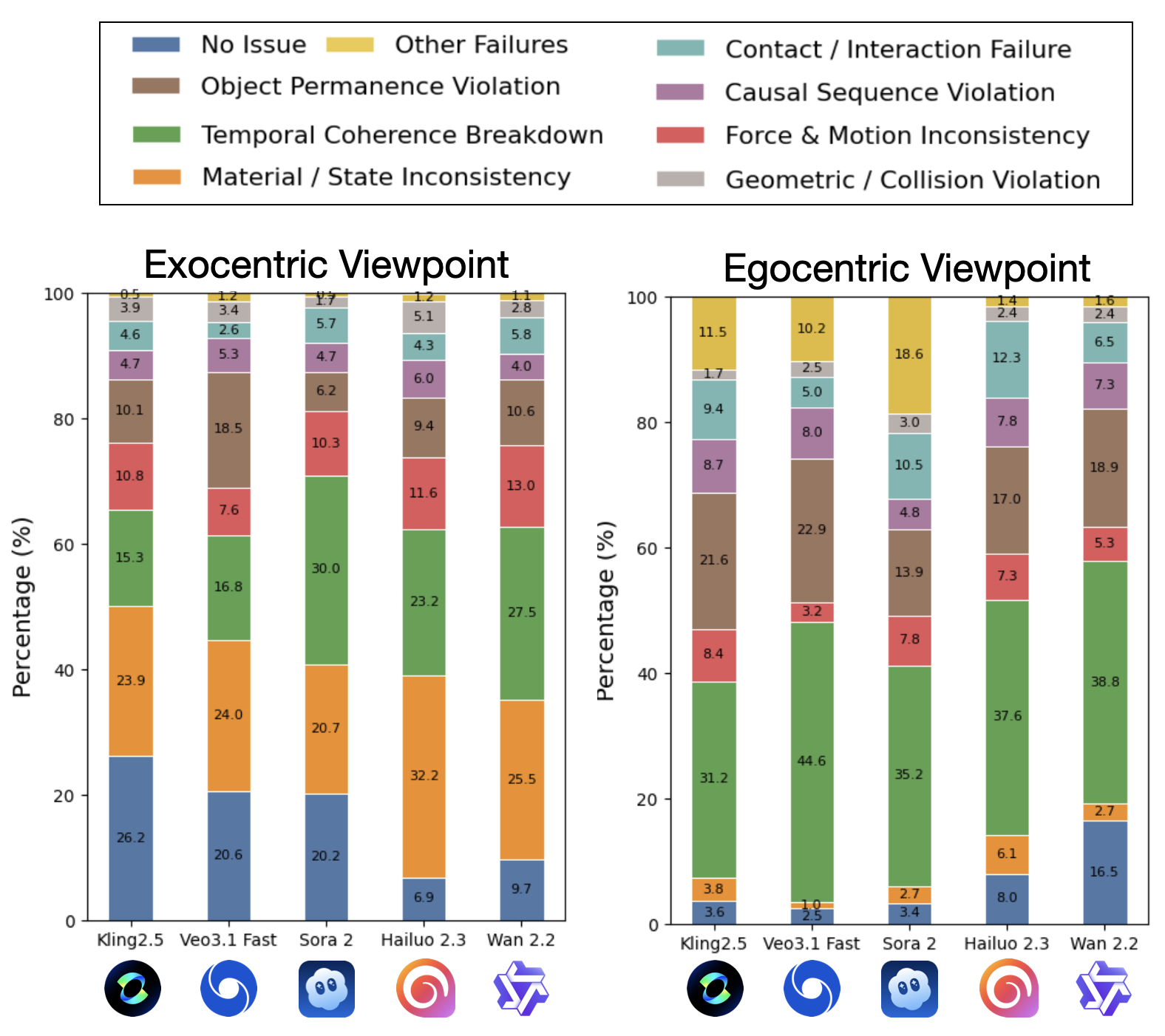}
    \caption{Distribution of physical glitch categories across models for exocentric (left) and egocentric (right) viewpoints. Stacked bars show the percentage of videos exhibiting different types of physical failures or no issue. Temporal coherence breakdown and material/state inconsistency are the most common failure modes in the exocentric view, while temporal coherence breakdown and object permanence violations dominate in the egocentric view.}
    \label{fig:stacked_bar}
\end{figure}

\subsubsection{Diagnosing Video Generation Models}
In \cref{fig:stacked_bar}, we analyze the distribution, frequency, and severity of physical glitches across models to diagnose failure modes in leading video generation models. In the exocentric setting, the most common failures are \textit{temporal coherence breakdown} and \textit{material/state inconsistency}. In the egocentric setting, \textit{temporal coherence breakdown} and \textit{object permanence violations} dominate. Overall, 83.3\% of exocentric and 93.5\% of egocentric videos contain at least one human-identified glitch. We also observe differences in the fraction of videos without observable physical issues: Kling 2.5 has the most glitch-free videos in the exocentric setting, while surprisingly, Wan 2.2, which is an open-source model, has the most glitch-free videos in the egocentric setting. We hypothesize that many commercial video models prioritize visual aesthetics and cinematic quality, which benefits third-person scenes but may reduce physical consistency in egocentric views where stable object dynamics are more critical. \cref{tab:glitch_metrics} further quantifies these differences using two metrics: \textit{glitch density}, defined as the average number of glitches per video, and \textit{glitch severity}, defined as the average severity score assigned by expert annotators on a 1–5 scale (higher indicates more severe violations). In the exocentric view, Kling 2.5 shows the lowest glitch density and severity among all models. In contrast, Sora 2 exhibits the lowest glitch density and severity in the egocentric view. Across models, glitch density ranges from roughly 1.15 to 1.92 glitches per video, indicating that multiple violations often occur within a single clip. Similarly, glitch severity remains consistently high, with average scores between 2.69 and 3.86, suggesting that many failures correspond to substantial physical inconsistencies. These results indicate that current video generation models still struggle to faithfully model physical dynamics, particularly in maintaining temporal consistency and preserving object continuity. 

\section{Conclusion}
This work presents the first large-scale, human-centered evaluation of physical realism in video generation. We show that outputs from state-of-the-art video generation models frequently violate basic physical principles: 83.3\% of exocentric and 93.5\% of egocentric videos contain human-identifiable glitches. Untrained human viewers reliably detect these failures, while current MLLM critics largely miss them, especially in egocentric settings, revealing a clear gap between human perception and automated evaluation. To support research in this direction, we introduce {Physion-Eval}, a benchmark containing 10,990 expert reasoning traces across 12,718 generated videos from five state-of-the-art models. The dataset provides temporally localized physical glitch annotations and natural-language explanations spanning 22 fine-grained physical phenomena across both egocentric and exocentric viewpoints. Beyond establishing a benchmark for perceptual physical realism evaluation, we hope Physion-Eval will enable:
\begin{enumerate}
\item \textit{Reasoning-driven diagnostics for video generation:}
Temporal reasoning traces enable precise identification of when and why physical violations occur.

\item \textit{Physically grounded video critics:}
The annotations enable training multimodal critics that detect, localize, and explain physical inconsistencies in generated videos.

\item \textit{Video generation with improved physical realism:}
Structured failure signals enable closed-loop, self-improving video generation systems, where models iteratively generate, diagnose, and refine outputs, aligning generation with the underlying laws of physical reality.
\end{enumerate}

\vspace{1pt}
\noindent\textbf{Limitations} \ Our evaluation mostly focuses on scenarios with a single dominant physical interaction and may not fully reflect complex multiphysics settings in the wild. Moreover, it relies on visually observable cues, so latent quantities (e.g., force, energy, entropy) are only indirectly inferred. Despite detailed guidelines and expert review, we expect some degree of annotation noise, as judgments of perceptual physical realism are inherently subjective.
{
    \small
    \bibliographystyle{ieeenat_fullname}
    \bibliography{main}
}

\clearpage
\setcounter{page}{1}
\maketitlesupplementary

\section{Task Definition}\label{defition} Physical realism evaluation in generated videos differs fundamentally from deepfake detection~\cite{rana2022deepfake}. Deepfake detection aims to determine whether a video is synthetic or not, typically relying on statistical artifacts or traces unique to each generative model~\cite{shen2025authguard}. In contrast, physical realism evaluation in generated videos asks whether the depicted semantics and dynamics obey real-world physical constraints. The task is therefore origin-agnostic: a synthetic video may be physically realistic, while a real video can exhibit implausible dynamics if it is manipulated. For example, temporal reversal (e.g., water refreezing into ice) or unrealistic speed changes can produce physically implausible behavior even in originally real footage. Thus, physical realism evaluation measures the perceptual plausibility of physical interactions rather than the authenticity of the media source. 

This distinction becomes increasingly important as modern video generation systems aim to simulate dynamic environments and complex interactions. In such settings, the key question is not merely whether the media is synthetic or real, but whether the depicted processes behave in a physically plausible manner that brings about immersive experience to the viewers. Physical realism evaluation therefore provides a more direct measure of the readiness of video generation systems for applications such as physical AI simulation, embodied agent training, and cinematic content production, where realistic motion, interaction, and temporal coherence are essential.

\section{Exocentric Video Curation}\label{sec:wisa_cleaning}
The exocentric videos are constructed from WISA-80K~\cite{wang2025wisa}, a large-scale video dataset covering 17 fundamental physical phenomena across dynamics, thermodynamics, and optics. Each video is paired with an original caption provided by the WISA dataset and a human-assigned physics category label. To ensure data quality and consistency for evaluation, we further curate the dataset by identifying and removing low-quality samples. During inspection, we observe that approximately 25\% of the original WISA videos exhibit quality issues that may affect their suitability for our benchmark use. We perform a detailed cleaning and quality-control process to retain videos with clear temporal structure, well-defined physical phenomena, and reliable annotations. Specifically, we manually:

\begin{enumerate}
\item Removed video clips containing multiple shots that break temporal continuity;
\item Excluded videos containing excessive or overlapping physical phenomena (e.g., many interacting objects or densely crowded scenes);
\item Discarded clips shorter than 2 seconds, temporally reversed videos (e.g., water refreezing into ice), computer-generated or animated content, and videos containing subtitles or text overlays;
\item Eliminated near-duplicate clips that correspond to different segments of the same original video; and
\item Removed video clips with incorrect physics labels (e.g., a car driving labeled as “gas motion” despite no visible gas flow).
\end{enumerate}

Moreover, WISA is originally heavily skewed toward a few categories, particularly reflection, liquid dynamics, and rigid-body motion. To obtain more balanced coverage across various physical phenomena, we further balance the dataset to achieve approximately near-uniform representation. All filtering steps were reviewed and validated by three physics PhD experts independently to avoid bias.

\section{Evaluation Prompts}\label{prompt}
\noindent\textbf{Prompt for Egocentric Video Caption.} \ Below is the instruction prompt used to generate captions for egocentric video segments curated from the EPIC-KITCHENS dataset.
\begin{tcolorbox}[colback=gray!10, colframe=gray!20, left=2mm, right=2mm, top=1mm, bottom=1mm, boxrule=0pt]
\small
You are an expert in video understanding and captioning for egocentric videos.  
Given a short {egocentric (first-person) video segment} corresponding to a {verb action label} \textbf{\textit{\{the corresponding action label\}}}, generate a concise caption describing the action occurring in the segment.

The caption must follow these requirements:
\begin{enumerate}
\item The caption should describe the scene explicitly {from a first-person viewpoint}.
\item The description must remain {objective and neutral}; do not use first-person pronouns such as {I}, {we}, or {my}.
\item Ensure the caption is {consistent with the provided verb label} and focuses on the {primary action and manipulated object(s)}.
\item Describe only {visually observable events} in the video and avoid speculation about intentions or unseen actions.
\item The caption should be {a single concise sentence}.
\end{enumerate}
The output must follow exactly the format specified below:
\begin{verbatim}
{"action_label": "text",
"caption": "text"}
\end{verbatim}
Return only the JSON object and no additional text.

\end{tcolorbox}

\noindent\textbf{Prompt for Physical Intensity and Dynamics Estimation.} \ Below is the instruction prompt used for generating physical intensity and dynamics in generated videos (to be discussed in Appendix~\cref{sec:physcial_dependency}).

\begin{tcolorbox}[colback=gray!10, colframe=gray!20, left=2mm, right=2mm, top=1mm, bottom=1mm, boxrule=0pt]
\small
You are an expert in physics and video understanding.  
Given a short video segment depicting a scene of \textbf{\textit{\{physical category\}}},  
analyze the sequence to infer its underlying physical properties. Your assessment should rely solely on visually observable evidence and reflect physically consistent reasoning.  
Evaluate the following two dimensions: 
\begin{enumerate}
\item \textbf{Intensity (scale 0–2)} -- Estimate the relative magnitude of the dominant physical process (0 = low, 1 = medium, 2 = high), calibrated with respect to the typical intensity range of \textbf{\textit{\{the given physical category\}}}. Base your rating on visually observable cues such as the scale of motion, degree of deformation, apparent momentum exchange, or energy release relative to what is normally expected for this phenomenon.
\item \textbf{Dynamics (scale 0–2)} -- Estimate the temporal activity of the process (0 = quasi-static or slow, 1 = medium, 2 = rapid or highly dynamic), calibrated with respect to the typical temporal behavior of \textbf{\textit{\{the given physical category\}}}. Base your rating on observable evidence such as motion frequency, rate of change, temporal coherence, and propagation speed relative to what is normally expected for this phenomenon.
\end{enumerate}

The output must follow exactly the format specified below:
\begin{verbatim}
{"intensity_score": int,
"intensity_reason": "text",
"dynamics_score": int,
"dynamics_reason": "text"
}
\end{verbatim}
Return only the JSON object and no additional text.

\end{tcolorbox}

\noindent\textbf{Prompt for MLLM Physical Glitch Detection.} \ Below is the instruction prompt used to evaluate whether MLLM critics can assess the physical realism of AI-generated videos. The model responses are then aggregated to produce the statistics reported in~\cref{fig:teaser}.

\begin{tcolorbox}[colback=gray!10, colframe=gray!20, left=2mm, right=2mm, top=1mm, bottom=1mm, boxrule=0pt]
\small
Does the behavior in this video appear physically {realistic} or {unrealistic}? Please answer based only on what you can observe in the video. Focus on motion, interactions, and whether the behavior follows real-world physical expectations.

\textbf{Output (strict JSON):}
\begin{verbatim}
{
  "label": "REALISTIC" | "UNREALISTIC",
  "confidence": int (0-100)
}
\end{verbatim}

Return only the JSON object.
\end{tcolorbox}

\noindent\textbf{Prompt for MLLM Physical Glitch Reasoning.} \ Below is the instruction prompt used to elicit detailed physical glitch reasoning from MLLM critics when evaluating generated videos, producing the responses shown in~\cref{fig:reasoning_trace} and~\cref{comparison_mllm_vs_human}.

\begin{tcolorbox}[colback=gray!10, colframe=gray!20, left=2mm, right=2mm, top=1mm, bottom=1mm, boxrule=0pt]
\small
You are a rigorous video forensics and physical-reasoning expert. Your task is to determine whether a given video is {PHYSICALLY\_REALISTIC} or {PHYSICALLY\_UNREALISTIC}. Physically unrealistic behavior is any visible motion, interaction, or state change that violates real-world physical constraints. If you detect any physically unrealistic behaviors, you must identify the specific timestamps where they occur. Base your judgment on physical consistency, temporal coherence, causal dynamics, and observable motion. Do not rely on aesthetic quality alone.

\textbf{Evaluation criteria:}
\begin{enumerate}
    \item {Physical consistency}: gravity, inertia, collisions, conservation laws
    \item {Temporal coherence}: object permanence, identity stability
    \item {Motion realism}: acceleration, contact dynamics, fluids, cloth
    \item {Causal structure}: actions produce physically plausible effects
    \item {Rendering artifacts}: artifacts affecting physical interpretation (e.g., over-smoothing, texture drift)
\end{enumerate}

\textbf{Guidelines:}
\begin{itemize}
    \item Use only observable visual evidence
    \item Do not infer hidden variables unless strongly supported
    \item Be conservative when evidence is ambiguous
    \item Include accurate timestamps (e.g., ``1.3s-3.1s") for any detected physical glitches
\end{itemize}

\textbf{Output (strict JSON):}
\begin{verbatim}
{
  "prediction": "PHYSICALLY_REALISTIC" | 
  "PHYSICALLY_UNREALISTIC",
  "confidence_score": int (0-100),
  "confidence_level": 1 | 2 | 3,
  "timestamps": ["0.0s-0.0s", ...] | null,
  "key_evidence": ["observation", ...],
  "reasoning": "2-4 sentences grounded in 
  physical evidence and causality"
}
\end{verbatim}

Return only the JSON object.
\end{tcolorbox}

\section{Ablation Studies}\label{ablation}
\subsection{Effect of Temporal Sampling on MLLM Performance} \label{temporal_sampling}
As shown in Appendix \cref{tab:temporal_sampling_effect}, increasing the temporal frame rate or the number of sampled frames per video produces only minimal and non-monotonic changes in Youden’s J statistic across both exocentric and egocentric settings. In all cases, LLM critics achieve $J < 13.7\%$ in the exocentric setting (compared to a human $J = 24.9\%$) and $J < 5.6\%$ in the egocentric setting (compared to a human $J = 58.5\%$. Manual inspection confirms that many physical glitches are obvious to human observers, indicating a genuine failure of the critic models rather than ambiguous cases. Notably, denser temporal sampling can sometimes even degrade $J$, as observed for both Gemini 3.0 Pro and GPT-5.2 in the exocentric setting.

\subsection{Effect of Thinking on MLLM Performance} \label{thinking}
We further examine whether enabling explicit thinking improves MLLM critics’ ability to detect and reason about physical glitches in generated videos by comparing Claude Opus 4.5 and GPT 5.2 with and without thinking\footnote{When thinking is enabled, Opus 4.5 uses a 2{,}000-token thinking budget, while GPT 5.2 employs a high reasoning setting.}. As shown in Appendix \cref{tab:thinking_vs_not}, enabling additional reasoning has a negligible impact, with a maximum $\Delta J$ of less than 2.0\%. One possible explanation is that the reasoning process largely operates in the language space. If the visual encoder fails to capture the fine-grained and often transient visual cues required to detect physical glitches, additional reasoning alone may provide limited benefit.

\begin{table}[t!]
\centering
\small
\footnotesize
\caption{
Effect of temporal sampling on MLLM critic performance measured by J-statistic, \% ($\uparrow$) in exocentric and egocentric views.}
\begin{tabular}{lc|c|c}
\toprule
\multirow{2}{*}{\textbf{Critic}} & \multirow{2}{*}{\textbf{Sampling Rate}}
& \multicolumn{1}{c|}{\textbf{Exocentric}}
& \multicolumn{1}{c}{\textbf{Egocentric}}
\\
 && {\tiny$G=\text{Kling~2.5}$} & {\tiny$G=\text{Veo~3.1 Fast}$}\\
\midrule
\textcolor{gray}{Human (reference)} & \textcolor{gray}{-} & \textcolor{gray}{24.9} & \textcolor{gray}{58.5} \\
\midrule
\multirow{3}{*}{Gemini~3.0~Pro} & 1 FPS & \textbf{13.7} & 3.7 \\
& 5 FPS  & 10.5 & \textbf{5.6} \\
 & 10 FPS  & 8.1 & 4.1\\
\midrule
\multirow{3}{*}{GPT~5.2} & 12 frames      &  \textbf{3.2} &  1.5 \\
 & 24 frames  & 2.1 & 2.0 \\
 & 48 frames  & 2.6 & \textbf{2.6} \\
\bottomrule
\end{tabular}
\label{tab:temporal_sampling_effect}
\end{table}

\begin{table}[t!]
\centering
\setlength{\tabcolsep}{4pt}
\renewcommand{\arraystretch}{1}
\setlength{\extrarowheight}{0pt}
\footnotesize
\caption{Critic performance with and without explicit reasoning when evaluating videos from Kling 2.5 (K) and Veo 3.1 Fast (V), measured by J-statistic, \% ($\uparrow$) in exocentric and egocentric views.}
\label{tab:thinking_vs_not}
\begin{tabular}{lc|c|c|c|c}
\toprule
\multirow{2}{*}{\textbf{Critic}} & \multirow{2}{*}{\textbf{Thinking}}
& \multicolumn{2}{c|}{\textbf{Exocentric}}
& \multicolumn{2}{c}{\textbf{Egocentric}} \\
& 
& \multicolumn{1}{c}{\tiny{Kling 2.5}}
& \multicolumn{1}{c|}{\tiny{Veo 3.1 Fast}}
& \multicolumn{1}{c}{\tiny{Kling 2.5}}
& \multicolumn{1}{c}{\tiny{Veo 3.1 Fast}} \\
\midrule
\textcolor{gray}{Human (reference)} & \textcolor{gray}{-} & \textcolor{gray}{24.9} & \textcolor{gray}{35.5} & \textcolor{gray}{48.4} & \textcolor{gray}{58.5}  \\
\midrule
\multirow{3}{*}{Claude Opus 4.5}
& No
& 4.1 
& 0.0
& 2.5 
& 2.6 \\
& Yes
& 4.6 \tiny{(+0.5)}
& 2.0 \tiny{(+2.0)}
& 4.3 \tiny{(+1.8)}
& 4.1 \tiny{(+1.5)}\\
\midrule

\multirow{3}{*}{OpenAI GPT~5.2}
& No
&  2.6
&  3.3
& 2.3
& 2.6 \\
& Yes
& 4.5 \tiny{(+1.9)}
& 3.5 \tiny{(+0.2)}
& 0.4 \tiny{(-1.9)}
&  3.5 \tiny{(+0.9)} \\
\bottomrule
\end{tabular}
\end{table}

\subsection{Effect of Physical Intensity and Dynamics on Generator and MLLM Critic Performance}\label{sec:physcial_dependency}

\begin{figure}[t!]
    \centering
    \includegraphics[width=1\linewidth]{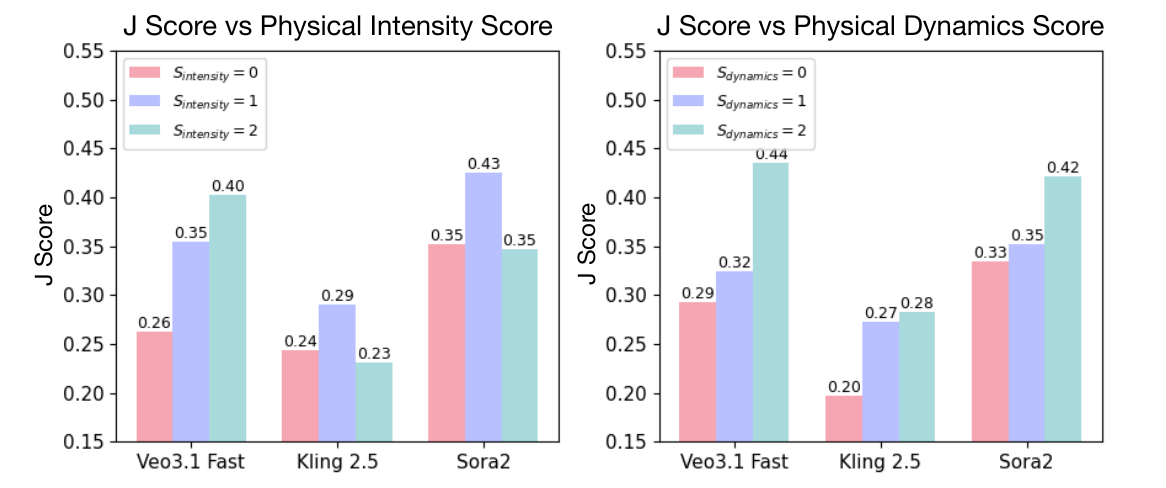}
    \caption{
        J values computed from untrained human evaluations for three video generation models in the exocentric setting, stratified by physical intensity and dynamics scores, denoted as $S_\text{intensity}$ and $S_\text{dynamics}$, respectively. Lower J statistic indicates stronger generator performance, corresponding to videos that are perceived as more physically realistic by ordinary human viewers. 
        \textcolor{pink}{Pink}, \textcolor{violet}{purple}, and \textcolor{cyan}{cyan} bars correspond to physical intensity or dynamics scores 0, 1, and 2, respectively.
    }
    \label{fig:intensity_dynamics_scores}
\end{figure}

\begin{figure}[t!]
    \centering
    \includegraphics[width=1\linewidth]{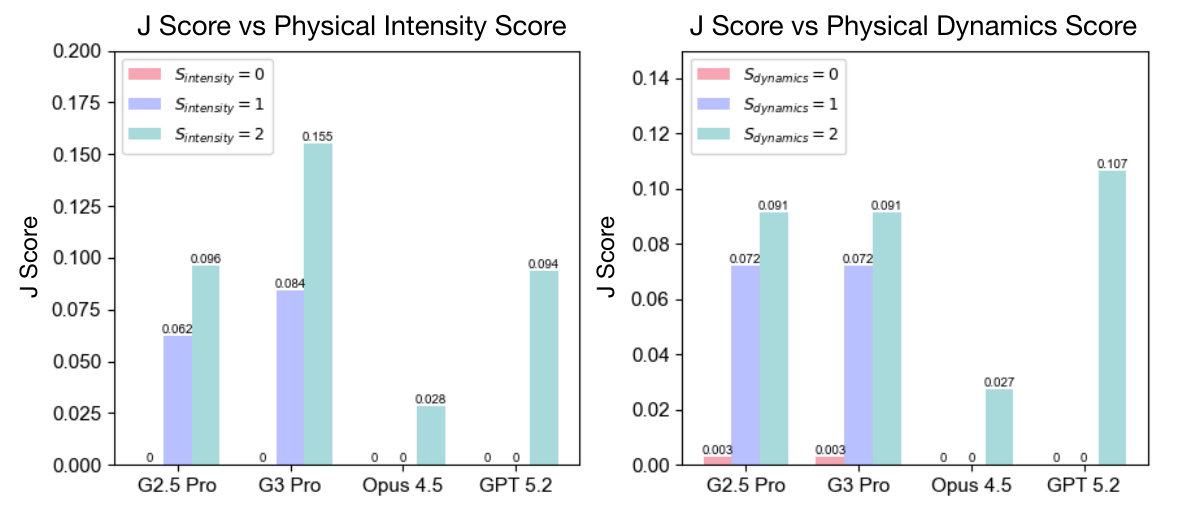}
    \caption{
        Comparison of J values obtained using different MLLM critics when evaluating videos generated by Veo3.1 Fast, stratified by three-level measures of physical intensity (left) and dynamics (right) in the exocentric setting. Higher J values indicate stronger critic ability to distinguish generated videos from real ones within a given physical regime. Zero J statistic values result in bars of zero height and are therefore not visible. G2.5 and G3 denote Gemini 2.5 and Gemini 3, respectively. \textcolor{pink}{Pink}, \textcolor{violet}{purple}, and \textcolor{cyan}{cyan} bars correspond to physical intensity or dynamics scores 0, 1, and 2, respectively.
    }
    \label{fig:J statistic_critic}
\end{figure}

\begin{figure*}[t!]
    \centering
    \includegraphics[width=1\linewidth]{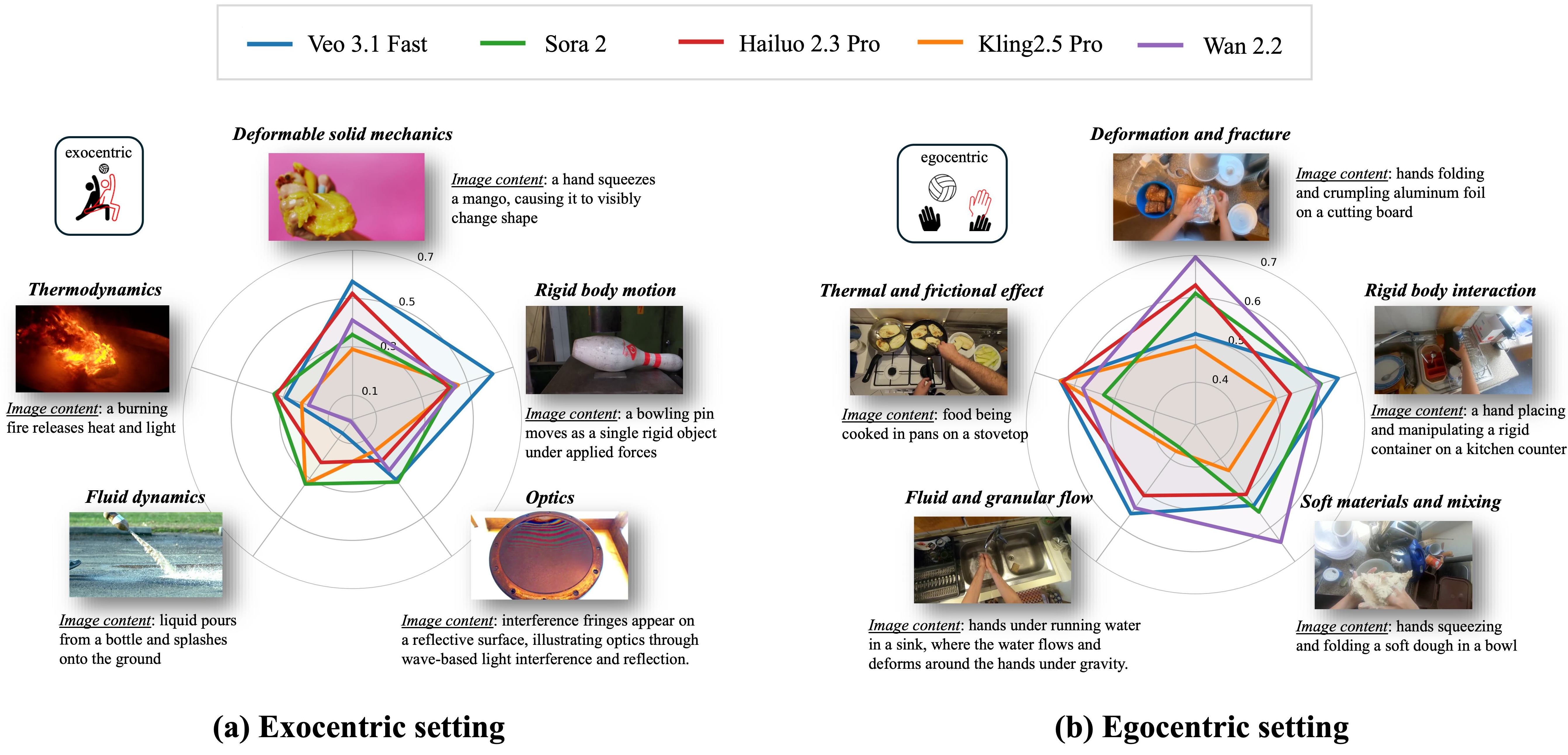}
    \caption{\textbf{Average physical realism judgments by untrained human observers across physics categories.} Radar plots show the J statistic computed from judgments by {ordinary human viewers} when evaluating videos generated by five video generation models in the {exocentric} (left) and {egocentric} (right) settings. Each axis corresponds to a physical phenomenon category, with representative frames and descriptions shown around the plots. The results reveal substantial variation across physics categories and models.
    }
    \label{fig:physical_category_comparison}
\end{figure*}

\noindent \textbf{Physical Metadata Extraction} \ To support a principled analysis across different physical categories, we extract high-level physical metadata along two core dimensions -- \textit{\textbf{intensity}} and \textit{\textbf{dynamics}} -- which characterize, respectively, the magnitude and temporal evolution of the physical processes of interest. Specifically, intensity reflects the dimensionless strength of physical interaction, ranging from low (gentle or near-static interactions such as slow placement or light contact), to medium (moderate force), to high (strong forces including collisions, rapid deformation, or splashing), and dynamics describes how rapidly physical states change over time, with low dynamics indicating slow or smooth motion, medium dynamics corresponding to moderate temporal variation, and high dynamics involving fast, abrupt, or highly transient motion. These metadata are inferred using Gemini 2.5 Pro~\cite{google2025gemini25} with the prompt provided in Appendix~\cref{prompt}. To validate the reliability of the LLM-estimated physical signals, we asked three PhD-level physics researchers to independently estimate intensity and dynamics using the same instructions given to Gemini. As shown in Appendix~\cref{tab:metadata_correlation}, across 335 exocentric videos (approximately 20 per category), their estimates show strong Pearson correlation ($>0.7$) with Gemini 2.5 Pro-extracted metadata, suggesting the estimates are reasonably accurate.

\vspace{1pt}
\noindent\textbf{Effect of Physical Intensity and Dynamics on Generator Performance.} Appendix~\cref{fig:intensity_dynamics_scores} shows the dependence of J statistic on three-point measures of physical intensity and dynamics for different video generators in the exocentric setting. Across all video generators, J statistic consistently increases with higher dynamics levels, suggesting that modeling temporally dynamic physical behavior remains comparatively challenging for today's video generation models. In contrast, the dependence on intensity is less consistent across models. For example, Veo3.1 Fast shows a monotonic increase, while Kling~2.5 and Sora~2 exhibit non-monotonic trends. Overall, judgments of physical realism by untrained human observers appear to be more strongly influenced by the dynamics of the physical processes than by their intensity in generated videos.

\vspace{1pt}
\noindent\textbf{Effect of Physical Intensity and Dynamics on Critic Performance.} Appendix~\cref{fig:J statistic_critic} shows J statistic values for different MLLM-based critics evaluating videos generated by Veo 3.1 Fast in the exocentric setting, stratified by increasing levels of physical intensity and dynamics. As shown, both Opus 4.5 and GPT 5.2 show little to no ability to assess physical realism at low and intermediate intensity and dynamics levels (scores $=0$ or $1$). At higher intensity and dynamics levels, J statistic increases across critics, indicating that some sensitivity to more physically demanding regimes emerges. We hypothesize that this is because current generators tend to produce more visibly implausible outputs in physics-intensive regimes, as also reflected in Appendix \cref{fig:intensity_dynamics_scores}, which may in turn make physical inconsistencies easier for critics to identify. However, despite this modest sensitivity, MLLMs remain substantially inferior to human performance overall.

\begin{figure*}[t!]
    \centering
    \includegraphics[width=0.9\linewidth]{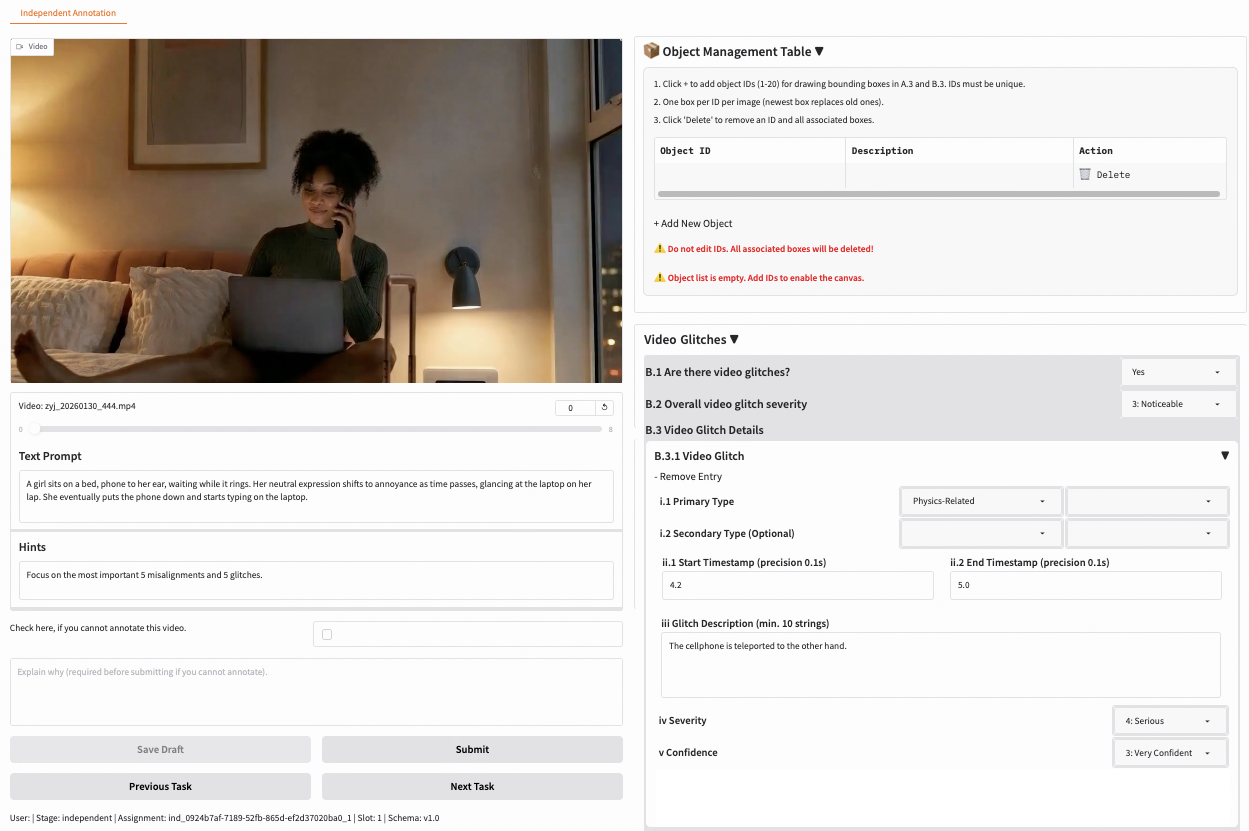}
    \caption{Expert Annotation Interface for Physically Grounded Video Evaluation.
    }
    \label{fig:interface}
\end{figure*}

\begin{table}
\centering
\caption{Pearson correlation (PCC) and 95\% confidence intervals between expert-annotated and Gemini 2.5 Pro–extracted physical metadata, with PCC $>0.7$ indicating strong correlation.}
\label{tab:metadata_correlation}
\scriptsize
\resizebox{\columnwidth}{!}{
\begin{tabular}{l|cc|cc}
\toprule
\multirow{2}{*}{{\centering\textbf{Metadata}}} & 
\multicolumn{2}{c|}{\centering\textbf{Intensity Score}} & 
\multicolumn{2}{c}{\centering\textbf{Dynamics Score}} \\
\cmidrule{2-3}\cmidrule{4-5}
& {PCC} & {95\% CI} & {PCC} & {95\% CI} \\
\midrule
\textbf{Value} & 0.724 & [0.657, 0.779] & 0.738 & [0.673, 0.790]  \\
\bottomrule
\end{tabular}
}
\end{table}

\vspace{1pt}
\noindent{\textbf{Evaluation of Generator Performance Across Physical Categories.}}
\ Appendix~\cref{fig:physical_category_comparison} shows that the ability of untrained human observers to judge physical realism varies substantially across physics categories. In the exocentric setting, observers are generally more sensitive to violations in categories involving rigid-body motion and deformable solids, where object motion and shape changes provide clear visual cues. In contrast, categories such as optics or thermodynamics tend to yield lower detection performance, likely because the relevant physical processes are less visually salient or require more expert domain knowledge to assess. A related pattern appears in the egocentric setting. Detection is highest for deformation and fracture and soft materials and mixing, where violations produce visible shape changes, and lower for thermal or frictional effects and fluid or granular flow, where the relevant dynamics are less directly observable. Model differences remain visible in this setting, with Kling~2.5 showing lower J statistics, indicating that its outputs are comparatively harder for untrained observers to identify as physically unrealistic. Overall, the results highlight two key observations. First, the detectability of physical inconsistencies is dependent on the type of physical phenomenon depicted. Second, despite being untrained, human observers can reliably detect physically implausible behavior in many scenarios. This suggests that perceptual cues are often sufficient to reveal violations of everyday physical expectations.

\begin{figure*}[t]
    \centering
    \includegraphics[width=\linewidth]{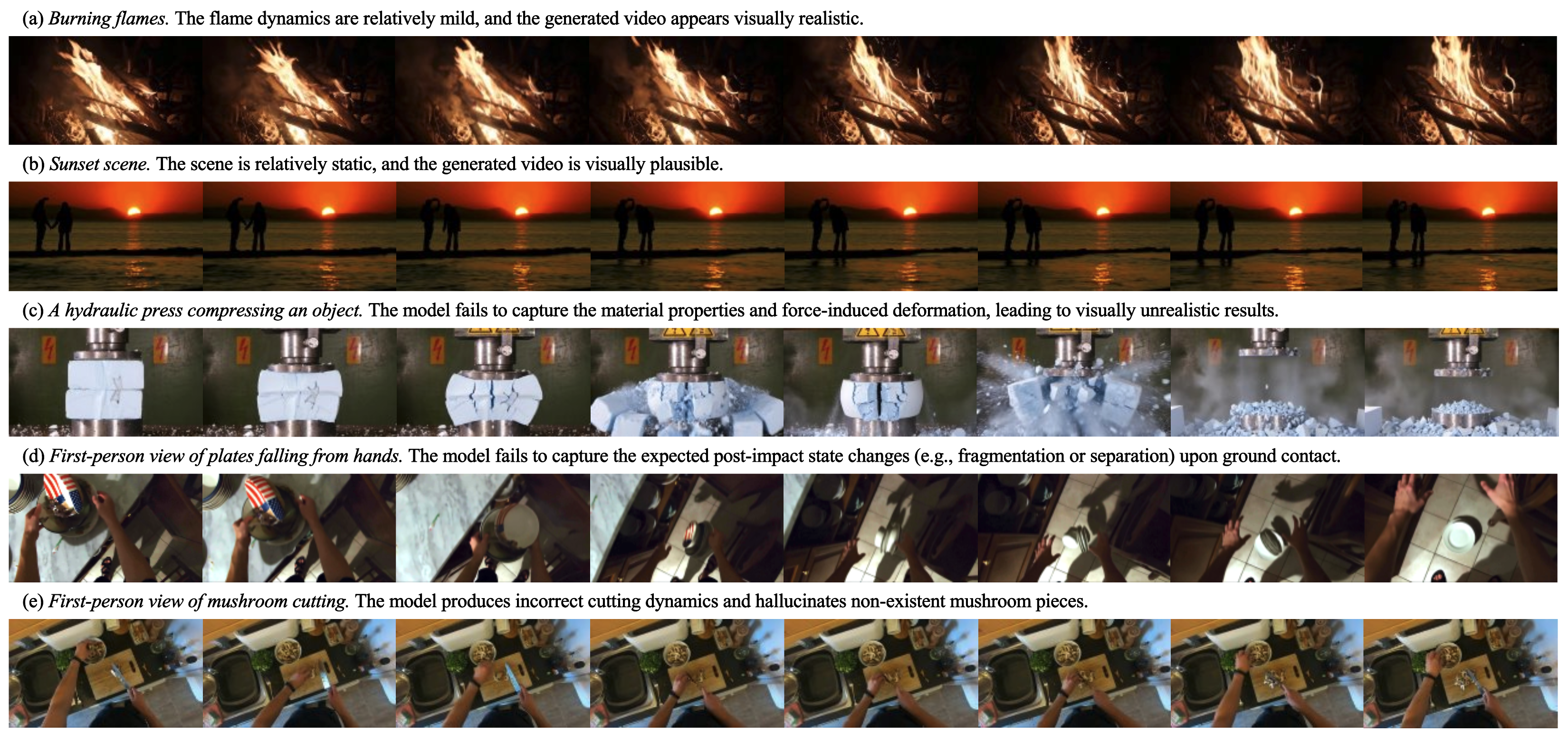}
    \caption{The first two videos (\textit{a} and \textit{b}) 
    show that the model generates visually plausible results for physical processes with low dynamics. In contrast, the latter three videos reveal unrealistic artifacts when the processes involve irreversible state changes and entropy-increasing dynamics (\textit{c}) 
    or abrupt and complex system-level transitions 
    (\textit{d} and \textit{e})
    .}
    \label{fig:four_examples}
\end{figure*}

\section{Expert Annotation Guidelines}\label{guideline}

Expert annotators evaluate short video clips (5--8 seconds, with audio) to identify perceptual violations of physical realism. Annotators are expected to apply structured reasoning and consistently use a predefined taxonomy to diagnose failures. For each video, annotators first determine whether it appears \textit{realistic} (consistent with real-world physical behavior) or \textit{unrealistic} (containing one or more violations of physical realism). For videos labeled as unrealistic, annotators identify one or more anomalies and, for each anomaly, provide the following: (1) \textit{taxonomy label(s)} describing the failure category; (2) a concise but precise \textit{description} grounded in observable evidence and physical reasoning; (3) \textit{temporal localization} indicating the approximate time range(s) where the anomaly occurs; and (4) a \textit{severity score} from 1 to 5 reflecting perceptual severity, where 1 denotes barely noticeable and 5 denotes severe and clearly implausible violations. If no glitch, severity score is labeled as 0. The failure taxonomy includes the following categories:
\begin{enumerate}
\item \textbf{Object Permanence Violation}: violations of object continuity over time, where an entity unexpectedly appears, disappears, duplicates, or changes identity without a physically plausible cause.
\item \textbf{Temporal Coherence Breakdown}: inconsistent rendering of a persistent entity across adjacent frames, where its visual attributes (e.g., texture, geometry details, or fine structure) change abruptly over time without physical cause, excluding cases of appearance, disappearance, or identity change.
\item\textbf{Material / State Inconsistency}: implausible material properties or state transitions, such as liquids behaving like solids or unnatural deformation. 
\item\textbf{Contact / Interaction Failure}: missing, incorrect, or physically implausible interactions between objects, including lack of contact response or hovering. 
\item\textbf{Causal Sequence Violation}: violations of cause-effect relationships, such as actions occurring before their causes or delayed and inconsistent responses. 
\item\textbf{Force \& Motion Inconsistency}: violations of basic physical dynamics, including gravity, inertia, acceleration, or momentum. 
\item \textbf{Geometric / Collision Violation}: physically impossible geometry or collisions, such as interpenetration or inconsistent object structure. 
\item \textbf{Other Failures}: anomalies that do not fit any of the predefined categories but still constitute perceptual violations of physical realism.
\end{enumerate}

Multiple anomalies may be annotated within a single video, each with independent labels and supporting evidence. Annotators are asked to avoid vague or purely subjective descriptions.

\begin{figure*}[t!]
    \centering
    \includegraphics[width=0.92\linewidth]{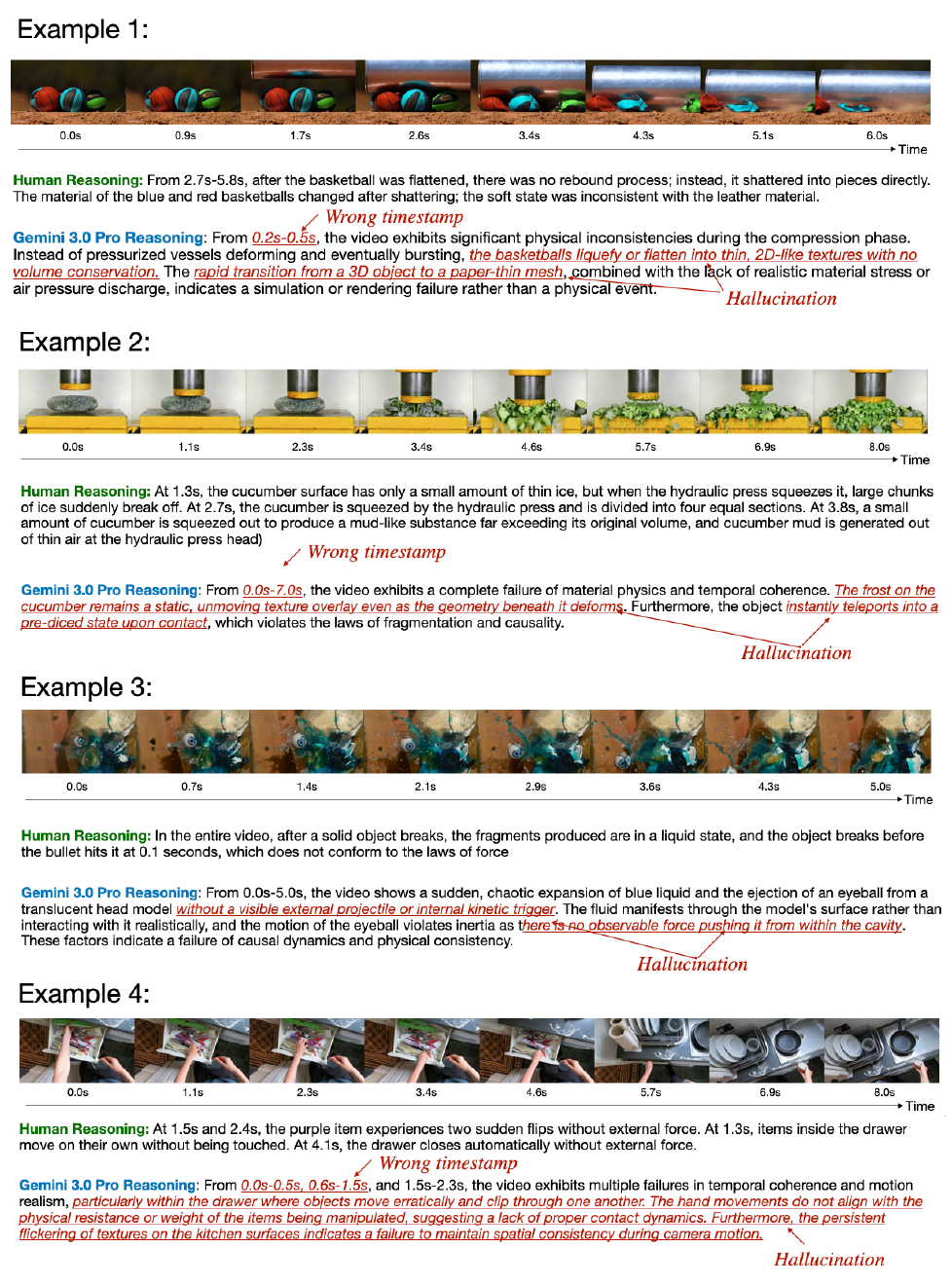}
    \caption{Comparison of expert human and MLLM reasoning on physical realism.}
    \label{fig:MLLM_reasoning_vs_human_reasoning}
\end{figure*}

\section{Qualitative Analysis of When Physical Realism Failures Tend to Emerge}\label{sec:anedocexample}
Visual inspection reveals consistent failure patterns in physically demanding scenarios, in line with the trends in Appendix \cref{fig:intensity_dynamics_scores}. As shown in Appendix \cref{fig:four_examples}, models perform well in low-intensity scenarios, such as viewing a static object from slowly changing angles or in approximately isolated settings like burning flames (\cref{fig:four_examples}a) and sunset scenes (\cref{fig:four_examples}b), where plausibility mainly depends on coarse temporal patterns and simple kinematics. In contrast, physical realism degrades sharply once scenes require non-trivial interactions, irreversible state changes, or complex system-level transitions. As shown in the hydraulic press example (\cref{fig:four_examples}c), the model fails to capture material properties and force-driven deformation: the object compresses in visually inconsistent ways, fragments unrealistically, and appears to partially recover despite the irreversible nature of the process. Similarly, in \cref{fig:four_examples}d-e, the generated video does not faithfully represent post-impact outcomes of plate dropping and food cutting, producing intact objects and unnatural merging. These failures concentrate in contact-rich and irreversible scenarios, such as collisions, material deformation, and entropy-increasing dynamics, where realism depends on enforcing conservation laws and causal state transitions. This pattern suggests that current video generation models rely on surface-level visual correlations rather than compositional physical understanding, leading to incoherent states in multi-object interactions and high-impact events which manifest as physical glitches.

\section{Comparison between MLLM Critic Reasoning and Human Reasoning}\label{comparison_mllm_vs_human}
The examples in \cref{fig:MLLM_reasoning_vs_human_reasoning} demonstrate that MLLM critics frequently produce explanations that are not supported by the visual evidence. In each case, the model refers to specific mechanisms, such as material deformation, texture overlays, invisible projectiles, or internal forces, that are not grounded in the visuals. In Example 2, it describes static frost textures and object “teleportation,” neither of which can be directly verified from the visual sequence. Similarly, in Example 3, the model hypothesizes an unseen internal force ejecting the eyeball. These hallucinated explanations are systematically paired with incorrect temporal grounding. The model often assigns failure intervals that do not align with when the physical inconsistency becomes visible (e.g., early timestamps that precede the onset of deformation or interaction). This suggests that, rather than tracking state changes over time, the model relies on coarse or preemptive judgments and then retrofits an explanation. In contrast, human annotators anchor their reasoning in observable state transitions, such as changes in material volume, fragmentation behavior, or object motion under force, and localize these failures to precise temporal segments. The discrepancy indicates that current MLLM critics do not reliably ground their reasoning in the visual evidence, and instead generate plausible-sounding but unsupported causal narratives.

\end{document}